\documentclass[letterpaper]{article} 
\usepackage{aaai23}  
\usepackage{times}  
\usepackage{helvet} 
\usepackage{courier} 
\usepackage[hyphens]{url}  
\usepackage{graphicx} 
\usepackage{natbib}  
\usepackage{caption} 
\usepackage{booktabs,multirow,adjustbox,diagbox,threeparttable}
\usepackage[utf8]{inputenc} 
\usepackage[T1]{fontenc}    
\usepackage{microtype}      
\usepackage{xcolor}         
\usepackage{xspace}
\usepackage{bbding}
\usepackage{amsmath,amssymb,amsfonts,dsfont,pifont,bm,bbm,mathrsfs,mathtools,nicefrac}
\usepackage{wrapfig}
\definecolor{citeblue}{RGB}{48,111,186}
\usepackage[pagebackref=false,breaklinks=true,colorlinks=true,citecolor=citeblue,bookmarks=false]{hyperref}
\usepackage{cleveref}
\usepackage{hyperref}
\crefname{section}{Sec.}{Secs.}
\Crefname{section}{Section}{Sections}
\crefname{appendix}{Appendix}{Appendixes}
\Crefname{appendix}{Appendix}{Appendixes}
\crefname{table}{Tab.}{Tabs.}
\Crefname{table}{Table}{Tables}
\crefname{figure}{Fig.}{Figs.}
\Crefname{figure}{Figure}{Figures}
\crefname{equation}{Eq.}{Eqs.}
\Crefname{equation}{Equation}{Equations}
\makeatletter
\newcommand\figcaption{\def\@captype{figure}\caption}
\newcommand\tabcaption{\def\@captype{table}\caption}
\makeatother

\usepackage{xcolor}         
\usepackage{algorithm}
\usepackage{algorithmic}
\usepackage[title]{appendix}

\usepackage{newfloat}
\usepackage{listings}
\DeclareCaptionStyle{ruled}{labelfont=normalfont,labelsep=colon,strut=off} 
\floatstyle{ruled}
\newfloat{listing}{tb}{lst}{}
\floatname{listing}{Listing}
\title{
Exploring Stochastic Autoregressive Image Modeling \\ 
for Visual Representation
}
\author{
    Yu Qi\textsuperscript{\rm 1}\equalcontrib\footnote{Intern at SenseTime Research},
    Fan Yang\textsuperscript{\rm 2}\equalcontrib,
    Yousong Zhu\textsuperscript{\rm 3},
    Yufei Liu\textsuperscript{\rm 1},
    Liwei Wu\textsuperscript{\rm 2},
    Rui Zhao\textsuperscript{\rm 2,4},
    Wei Li\textsuperscript{\rm 2}\thanks{Corresponding authors.}
}
\affiliations{
    \textsuperscript{\rm 1} Tsinghua University
    \textsuperscript{\rm 2} SenseTime Research
    \textsuperscript{\rm 3} Institute of Automation, Chinese Academy of Sciences \\
    \textsuperscript{\rm 4} Qing Yuan Research Institute, Shanghai Jiao Tong University, Shanghai, China

    qiy20@mails.tsinghua.edu.cn, 
    yousong.zhu@nlpr.ia.ac.cn,
    liuyufei@tsinghua.edu.cn\\
    \{yangfan1,wuliwei,zhaorui,liwei1\}@sensetime.com
    
}

\usepackage{bibentry}

\begin{document}

\maketitle

\begin{abstract}

 Autoregressive language modeling (ALM) has been successfully used in self-supervised pre-training in Natural language processing (NLP).
However, this paradigm has not achieved comparable results with other self-supervised approaches in computer vision (\textit{\textit{e.g.}}, contrastive learning, masked image modeling).
In this paper, we try to find the reason why autoregressive modeling does not work well on vision tasks.
To tackle this problem, we fully analyze the limitation of visual autoregressive methods and proposed a novel stochastic autoregressive image modeling (named SAIM) by the two simple designs.
First, we serialize the image into patches. Second, we employ the stochastic permutation strategy to generate an effective and robust image context which is critical for vision tasks.
%
To realize this task, we create a parallel encoder-decoder training process in which the encoder serves a similar role to the standard vision transformer focusing on learning the whole contextual information, and meanwhile the decoder predicts the content of the current position so that the encoder and decoder can reinforce each other.
%
%
Our method significantly improves the performance of autoregressive image modeling and achieves the best accuracy (83.9\%) on the vanilla ViT-Base model among methods using only ImageNet-1K data. Transfer performance in downstream tasks also shows that our model achieves competitive performance.
Code is available at \href{https://github.com/qiy20/SAIM}{https://github.com/qiy20/SAIM}.

\end{abstract}

\begin{figure*}[t]
    \centering
    \includegraphics[width=1\textwidth]{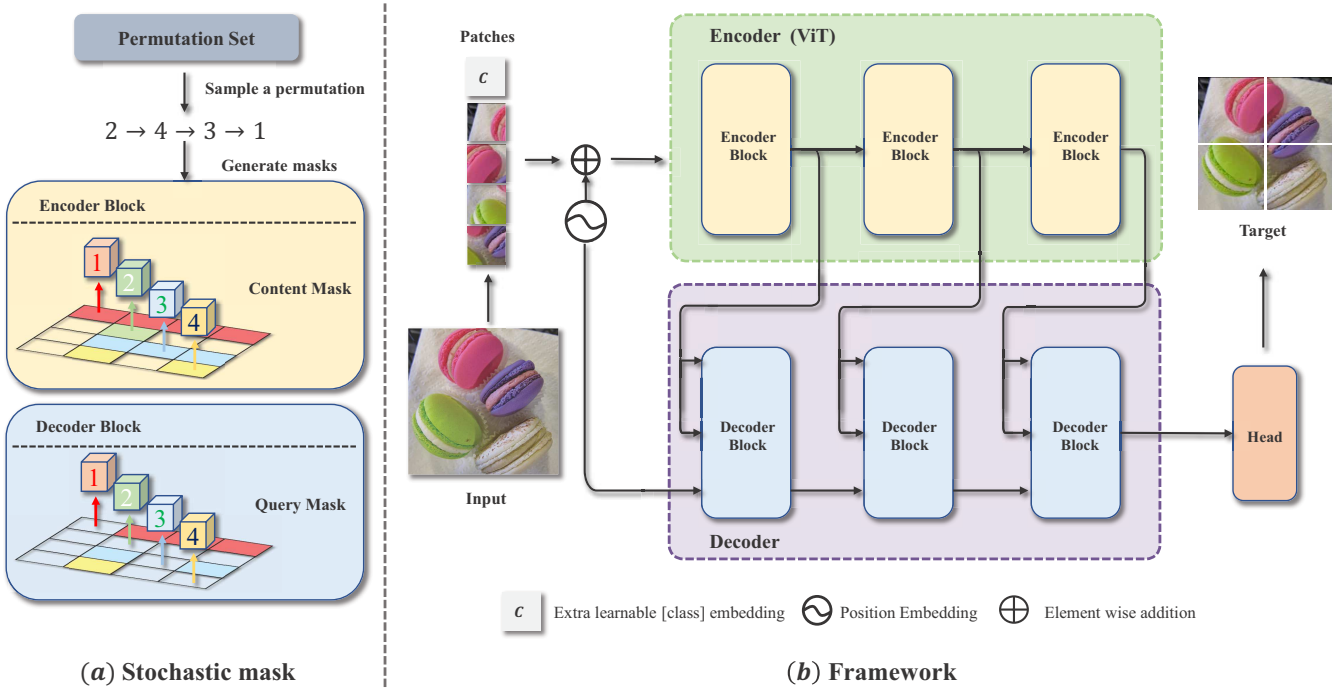}
    \caption{
        \textbf{Training pipeline of Stochastic Autoregressive Pretraining.}
        \textbf{Pre-training stage:} First, we randomly sample a permutation from the permutation set and generate two masks: content mask and query mask. 
        Second, we employ a parallel encoder-decoder for autoregressive prediction.
        The encoder focus on learning contextual information(only the visible positions on the content mask), 
        and the decoder reconstructs the original image from the latent representation along with position embeddings.
        \textbf{Fine-tune stage:} The parameters of the encoder are directly applied to the standard Vision Transformer (ViT) for downstream tasks.
        }
    \label{fig:pipeline}
\end{figure*}

\section{Introduction}\label{sec:Introduction}

Un-/Self-supervised representation learning has achieved great success in natural language processing (NLP)~\cite{yang2019xlnet, brown2020language, devlin2018bert, radford2018improving, radford2019language}.
Autoregressive language modeling (ALM) and masked language modeling (MLM) (\textit{e.g.}, GPT~\cite{radford2018improving, radford2019language, brown2020language} BERT~\cite{devlin2018bert}), are capable of training large-scale language models with human-like performances using billions of unlabeled training data.
Motivated by the success of BERT and the outstanding performance of Vision Transformers~\cite{dosovitskiy2020image}, recent works~\cite{he2021masked, bao2021beit, wei2021masked, xie2021simmim} introduce BERT-style pretraining by reconstructing the masked patches, which achieve an overall improvement in downstream tasks and greatly narrow the gap between vision and language.
OpenAI made an attempt to propose an influential work called iGPT~\cite{chen2020generative}, it learns representations by predicting pixels in raster order but the performances lag behind other self-supervised learning methods in both capacity and efficiency. 

We ask: \textit{what makes autoregressive modeling different between vision and language tasks?}
We try to answer this question from the following aspects: 

\begin{itemize}

    \item \noindent\textbf{Input signal.}
    Different from language which follows the fixed natural order, images are not sequential signals. The lack of a well-defined order is the main challenge of applying autoregressive methods to process images. Previous methods, such as PixelCNN~\cite{van2016pixel} and iGPT, just follow the raster order to generate pixels. Such order is perhaps optimal for image generation, but it may not be the best order for visual representation learning.
    Because when people look at an image, they first focus on the main object or the object they are interested in, which is randomly distributed in any position instead of fixed in the top left corner. So we propose to learn representations by predicting the object in stochastic order, which can take advantage of the richness and variety of visual signals.
    \item \noindent\textbf{Architecture}.
    In vision, convolutional networks have been the mainstream model until recently, but in the field of NLP, Transformer was dominant. iGPT firstly trains a sequence Transformer to auto-regressively predict pixels. But it can only work with low-resolution images because the computational complexity of self-attention is quadratic to image size, which prevents the capabilities and scaling of such approaches. So we introduce Vision Transformer as the encoder to solve this problem.
    As for the decoder, we design the parallel encoder-decoder architecture which is inspired by XLNET~\cite{yang2019xlnet} but doesn’t share the weights. 
    Benefit from our architecture, the encoder can focus on learning semantic representations without participating in pixel prediction.
    \item \noindent\textbf{Prediction target.} 
    For autoregressive image modeling, the prediction target is raw pixels containing a lot of "noise". This is in contrast to language, where the model predicts words that are high-level concepts generated by humans.
    We think that the prediction of low-level signals will overfit the high-frequency details and textures, which are not helpful for high-level recognition tasks.
    To overcome this problem we employ Gaussian smoothing to discourage the model from learning high-frequency details and encourage learning semantic information.
\end{itemize}
Based on the findings and analysis, we propose the stochastic autoregressive image modeling (SAIM), as shown in ~\cref{fig:pipeline}. 
First, we generate two masks from a random permutation to achieve stochastic autoregression, the content mask and the query mask will apply them to the encoder and the decoder respectively. With the masks, each query can only hold the preceding information and the position embedding of itself.
Second, we design the parallel encoder-decoder architecture.  The encoder focus on learning contextual information, and the decoder reconstructs the original image from the latent representation along with position embeddings.
Coupling these two designs enables us to realize stochastic sequence autoregression and achieve good performance.
In addition, applying Gaussian smooth to target images also improves our performance.
We pretrain the model on ImageNet-1K and then fine-tune on three downstream tasks. For ViT-base, it achieves 83.9\% top-1 fine-tuning accuracy on ImageNet-1K~\cite{russakovsky2015imagenet}, 49.4/43.9 box/mask mAP on COCO object detection~\cite{lin2014microsoft} and 47.8 mIoU on ADE20K semantic segmentation~\cite{zhou2019semantic}.
Experimental results indicate that SAIM consistently improves performance and achieves competitive performance with the state-of-the-art methods, it proves that the simple \textit{\textbf{Autoregressive Image Modeling}} is also an effective pretext task for visual representation learning. 

\section{Related work}\label{sec:related}

\textbf{Autoregressive language modeling (ALM)} and masked language modeling (MLM), \textit{e.g.}, GPT~\cite{radford2019language, radford2018improving, brown2020language} and BERT~\cite{devlin2018bert}, are two main self-supervised learning approaches in the field of NLP.
Given a sequence, BERT holds out a portion of the input sequence and tries to predict the missing tokens. GPT, on the hand, predicts all tokens following the left-to-right natural order of languages.
This series of works have achieved great success in the field of natural language processing. 
To leverage the best of both BERT and GPT, the stochastic autoregressive language modelings \textit{e.g.}, XLNet~\cite{yang2019xlnet} enables learning bidirectional contexts by maximizing the expected likelihood over all permutations of the factorization order. This strategy enables the ALM to acquire contextual information since the predicted targets can view all parts of the sequence during training.
We mainly take inspiration from the permutation-based autoregressive pretraining.


\textbf{Self-supervised learning (SSL)} aims to learn a general visual representation from unlabeled data and have good transfer performance in downstream tasks.
In the field of computer vision, researchers proposed a wide range of pretext tasks, such as image colorization~\cite{zhang2016colorful}, jigsaw puzzle~\cite{noroozi2016unsupervised}, image inpainting~\cite{pathak2016context}, rotation prediction~\cite{gidaris2018unsupervised}, and so on.
Such self-supervised pretraining has limited success, but has seen significant interest in this field.
Contrastive learning ~\cite{he2020momentum, chen2020big,caron2021emerging,chen2021empirical,chen2020simple,koch2015siamese,grill2020bootstrap} have dominated self-supervised pre-training in recent years, which makes the representations of positive pairs similar while pushing negative pairs away.
While, these methods need to learn representation from the object-centered datasets, and have poor transferability to dense prediction tasks (e.g. object detection and semantic segmentation).


\textbf{Masked image modeling (MIM)} introduces BERT-style pretraining into computer vision. ViT~\cite{dosovitskiy2020image} predicts the mean colors of masked patches. BEiT~\cite{bao2021beit} encodes masked patches with discrete variational autoencoder~\cite{ramesh2021zero} and uses the visual token as the prediction target. MAE~\cite{he2021masked} masks a high proportion of the input image and just predicts raw pixels. SimMIM~\cite{xie2021simmim} simply the pipeline of MIM methods. IBOT~\cite{zhou2021ibot} and data2vec~\cite{baevski2022data2vec} perform masked prediction with an online tokenizer. These methods achieve better transfer performance than supervised learning and contrastive learning. We observe that on the NLP side, GPT models also have shown to be powerful, while researchers pay more attention to BERT-style pre-training in computer vision. 

\textbf{Autoregressive image modeling (AIM)} is a classic approach in computer vision but located in a non-mainstream position for a long time.
PixelCNN~\cite{van2016conditional} and VQ-VAE~\cite{van2017neural} model the distribution of natural images and generate new images based on pre-trained models.
Pix2Seq~\cite{chen2021pix2seq} uses autoregressive modelling for object detection.
For representation learning, CPC~\cite{oord2018representation} predicts patches by learning an autoregressive model in the latent space.
The iGPT~\cite{chen2020generative} serializes pixels in raster order to make autoregressive predictions on low-resolution images.
The computational complexity of self-attention limits the extension of this method.
Recent advances in vision architectures, such as ViT, which serializes visual 2D data, provide an opportunity to apply similar large-scale pre-training in vision.
%
%
Our work follows this line, attempting to promote the development of AIM and bridge the gap between vision and natural language processing.


\section{Methods}\label{sec:method}


\subsection{Background}\label{sec:Background}
Self-supervised learning aims to learn a good visual representation from an unlabeled dataset. Autoregressive modeling can achieve this goal by modeling the distribution of natural signals, which is a landmark problem in SSL~\cite{van2016pixel}.

Speciﬁcally, given an unlabeled detaset $\mathcal D$ consisting of high dimensional data $\boldsymbol {x}=\left[x_{1},x_{2}, \cdots, x_{N}\right]$, we can pick a permutation $\boldsymbol {z}$ of the set $[1, N]$, and we use $z_i$ and $\boldsymbol z_{<i}$ to denote the {\it i}-th element and the first {\it i}-1 elements of the permutation. Autoregressive methods perform pretraining by maximizing the likelihood function:
\begin{equation}
\label{eq:ar}
\mathcal L= - \mathop {\mathbb E} \limits_{\boldsymbol x \sim \mathcal D} ~\sum_{i=1}^{N} \log p_{\theta}\left(x_{z_i} \mid \boldsymbol{x}_{{\boldsymbol z_{<i}}}\right)
\end{equation}
Where $\theta$ is the parameters of the autoregressive model. When working with images, ~\cite{chen2020generative} just flatten the pixels  in fixed raster order, e.g., $x_i$ donates a pixel and $z_i=i$ for $1\le i\le n$. But as we have analyzed before, there are two main limitations of this method: first, processing the pixel sequence is particularly time/space consuming~\cite{chen2020generative}; second, predicting in raster-order is not consistent with the human visual mechanisms.
\subsection{Stochastic autoregressive image modeling}\label{sec:Stochastic autoregressive image modeling}
Our proposed SAIM first serializes the image into patches and then predicts patches in stochastic order, which overcomes the above limitations.

\textbf{Image serialization}.
Following ViT~\cite{dosovitskiy2020image}, we first split the 2D image $\boldsymbol x\in \mathcal D$  into patches, and the image patches are flattened into vectors $\{ x_i \}_{i=1}^N$, where $\mathit N$ is the number of patches.
Then, the vectors are linearly projected to obtain patch embeddings $\boldsymbol Wx_i\in\mathbb R^{D}$. 
Finally, We add 2D sin-cos position embeddings $\boldsymbol E_{pos}=[e_1,e_2,\cdots,e_N]$ to patch embeddings, where $\boldsymbol E_{pos}\in \mathbb R^{N\times D}$.
So we get the initialized sequence $\boldsymbol s=[s_1,s_2,\cdots,s_N]=[\boldsymbol Wx_1,\boldsymbol Wx_2,\cdots,\boldsymbol Wx_N]+\boldsymbol E_{pos}$.

\textbf{Stochastic prediction}.
Instead of using a fixed permutation as in conventional autoregressive (AR) models.
Our approach borrows ideas from XLNet~\cite{yang2019xlnet}, using all possible permutations as the prediction order. 
Specifically, for a sequence of length $N$, there are $N!$ different orders to perform an auto-regressive model.
Let $\mathcal Z_N$ be the set of all possible permutations of the index set $\{1,2,\cdots,N\}$.
We predict the target patch depending on the preceding sequence $s_{\boldsymbol z_{<i}}$ and target position embedding $e_{z_i}$ and train our model by minimizing the mean squared error between the reconstruction and original image pixels:
\begin{equation}
\label{eq:objectibve function}
\mathcal L=\mathop {\mathbb E} \limits_{\boldsymbol x \sim \mathcal D} ~\mathop {\mathbb E} \limits_{\boldsymbol z \sim \mathcal Z_N} ~\sum_{i=1}^N || f_\theta(s_{\boldsymbol z_{<i}},~e_{z_i})-x_{z_i} ||^2
\end{equation}
where $\theta$ is the parameters of the model and  $f_\theta(s_{\boldsymbol z_{<i}},~e_{z_i})$ is the output of the model.
%
\subsection{Parallel encoder and decoder architecture}\label{sec:Parallel encoder and decoder architecture}
Inspired by~\cite{yang2019xlnet}, we don't permute the input sequence directly but rely on the two-stream self-attention with mask to implement SAIM. Based on the analysis in introduction chapter, the design of decoder is important for our method. So we design a parallel encoder-decoder architecture, in which the encoder doesn't share weights with the decoder. 
In the pre-training stage, the encoder focuses on learning contextual information(only the visible positions on the content mask), and the decoder reconstructs the original image from the latent representation along with position embeddings.
In the finetune stage, only encoder will be reserved and no masks will be applied.
%


\textbf{Mask generation}. For a sequence of length $\mathit N$, we first randomly generate a vector $\boldsymbol r=[r_1,r_2,\cdots,r_N]$, where $r_i$ follows a uniform distribution. So we can get the permutation $\boldsymbol z= \mathrm{argsort}(\boldsymbol r)$. The \textbf{content mask} can be generated as follows:
\begin{equation}
\label{eq:generate mask}
\mathrm{content\_mask}_{ij}=\left\{
\begin{aligned}
0,       &      & {r_i <      r_j}\\
1,       &      & {r_i \geq r_j}\\
\end{aligned}
\right.
\end{equation}
Where $i,j$ are the coordinate of attention matrix, $\mathrm{content\_mask}_{ij}=1$ represents the $i$-th token have access to the $j$-th token, $\mathrm{content\_mask}_{ij}=0$ just represents the opposite. The \textbf{query mask} is basically the same shape as the content mask, except that the diagonal is all zero.

\textbf{Encoder}.
The encoder has the same structure as the Vision Transformer, which consists of $M$ layers of self-attention blocks, and we apply a \textbf{content mask} to the self-attention blocks, which makes the current token only gather information from the preceding positions. 
Computationally, we define $h_i^{(m)}$ as the output of the $m$-th encoder layer, where $i$ is the token index. 
And we use the initialized sequence $\boldsymbol s$ as the input of the first encoder layer, i.e. $h_i^{(0)}=s_i$, the forward process of the encoder can be described as follows:
\begin{equation}
\begin{aligned}
h_{z_t}^{(m)}=&{\rm Attention}(
{\rm Q}=h_{z_t}^{(m-1)},{\rm KV}=h_{\boldsymbol z_{\le t}}^{(m-1)};\theta_{e}^{(m)})\\
&where~~1\le m \le M
\end{aligned}
\end{equation}
Where $\theta_{e}^{(m)}$ is the parameters of the $m$-th encoder layer. We omit the layer norm, MLP, and residual connection in the notation.

\textbf{Decoder}.
The decoder consists of $M$ layers of cross-attention blocks and a MLP layer.
The cross-attention blocks reconstruct the original signals,
and the MLP layer projects the signal to the initial dimension.
We also apply a \textbf{query mask} to the cross-attention blocks.

We can define $g_i^{(m)}$ as the output of the $m$-th decoder layer.  And we use the position embeddings $\boldsymbol E_{pos}$ as the input of the first decoder layer, i.e. $g_i^{(0)}=e_i$, the forward process of the decoder can be described as follows:
\begin{equation}
\begin{aligned}
g_{z_t}^{(m)}=&{\rm Attention}({\rm Q}=g_{z_t}^{(m-1)},{\rm KV}=h_{\boldsymbol z_{< t}}^{(m-1)};\theta_{d}^{(m)})\\
&where~~1\le m \le M
\end{aligned}
\end{equation}
\begin{equation}
\begin{aligned}
g_{z_t}^{(m)}=&{\rm MLP}(g_{z_t}^{(m-1)};\theta_{d}^{(m)})~~~~~~~~~~~~~~~~~~~~~~~~~~~~~~~~~~~~~~~~\\
&where~~m=M+1
\end{aligned}
\end{equation}
where $\theta_{d}^{(m)}$ is the parameters of the $m$-th decoder layer, which is different from  $\theta_{e}^{(m)}$.
Finally, we can use the output of the last decoder layer
$g_{z_t}^{(M+1)}$
to compute loss, e.g. $f_\theta(s_{\boldsymbol z_{<i}},~e_{z_i})=g_{z_t}^{(M+1)}$ in \cref{eq:objectibve function}.
\subsection{Gaussian smoothing application}\label{sec:Gaussian smoothing application}
Vision signals are raw and low-level, and high-frequency details and textures are not helpful for common recognition tasks.
To make the model focus on learning semantic information, we construct a two-dimensional Gaussian filter kernel to reduce the texture detail of the target.
The \cref{eq:objectibve function} can update as:
\begin{equation}
\mathcal L=\mathop {\mathbb E} \limits_{\mathbf x \sim \mathcal D} ~~\mathop {\mathbb E} \limits_{\mathbf z \sim \mathcal Z_N} ~~\sum_{i=1}^N||f_\theta(s_{\mathbf z_{<i}},~e_{z_i})-g_\xi (x_{z_i}) ||^2
\end{equation}


where $g_\xi$ denotes the Gaussian filter. This simple strategy works well in our method and the computation is negligible.

\section{Experiments}

\begin{table*}[ht]
\caption{\textbf{Comparison with previous results on ImageNet-1K.}
 $\ast$: iGPT-L contains 1.36 billion parameters, while others use ViT-base model. $\dag$: combine patch-wise modeling of ViT and raster ordering prediction of iGPT}
\begin{center}
\begin{tabular}{lcccc}
\toprule
\textbf{Method} & \textbf{Arch} &\textbf{Pretrain epochs} &\textbf{Linear} &\textbf{Fine-tune} \\ 
\midrule 
DeiT~\cite{touvron2021training} & ViT-B &300& - & 81.8 \\
\midrule 
MoCoV3~\cite{chen2021empirical} & ViT-B &300& 76.2 & 83.0 \\
DINO~\cite{caron2021emerging} & ViT-B &400& 77.3 & 83.3 \\
\midrule 
BEiT~\cite{bao2021beit} & ViT-B &800& - & 83.2 \\
MAE~\cite{he2021masked} & ViT-B &1600& 67.8 & 83.6 \\
simMIM~\cite{xie2021simmim} & ViT-B &1600& 56.7 & 83.8 \\
CAE~\cite{chen2022context} & ViT-B &800& 68.3 & 83.6 \\
\midrule 
$\rm{iGPT}^\ast$~\cite{chen2020generative}&iGPT-L& - & 65.2 &  72.6 \\
RandSAC~\cite{hua2022self}& ViT-B & 1600 & 68.3 &  83.0 \\
ViT-iGPT$^\dag$  & ViT-B & 300 & 20.4 & 82.7 \\
SAIM  & ViT-B & 300 & 58.5 & 83.6 \\
SAIM  & ViT-B & 800 & 62.5 &  \textbf{83.9} \\
 \bottomrule
\end{tabular}
\end{center}
\label{tabimage_v2}
\end{table*}

\subsection{Pretraing setup}
\label{sec:PretraingSetup}
\textbf{Dataset and models}. Our method is pretrained on the popular ImageNet-1k~\cite{russakovsky2015imagenet} dataset.
The dataset contains 1.28 million images from the training set of 1000 classes and 50,000 images from the validation set.
We only use the training set during self-supervised learning. 
Our default model architecture has the parallel encoder and decoder.
The encoder keeps the same structure as Vision Transformer~\cite{dosovitskiy2020image}, and the decoder consists of cross-attention blocks and a MLP layer. 
Visual Transformer will load our trained encoder weights for downstream task evaluation.


\textbf{Training configurations}. We use AdamW for
optimization and pretraining for 300/800 epochs with the batch size being 2048. We set the base learning rate as 2e-4, with cosine learning rate decay and a 30-epoch warmup, and set the weight decay as 0.05. We do not employ drop path and dropout.
A light data augmentation strategy is used: random resize cropping with a scale range of [0.67, 1] and an aspect ratio range of [3/4, 4/3], followed by random flipping and color normalization steps.

\subsection{Transfer learning on downstream vision tasks}
\label{sec:experiments}

\textbf{Image classification on ImageNet-1K}.
We conduct fine-tuning and linear probing experiments on ImageNet-1K image classification in ~\cref{tabimage_v2}.
The fine-tuning setting follows the common practice of supervised ViT~\cite{dosovitskiy2020image} training. Implementation details are in Supplemental.
Our model pretrained with 300 epochs achieves the same accuracy as MAE~\cite{he2021masked} pretrained with 1600 epochs, which indicates that our method converges faster in the pretraining stage. Our hypothesis here is that our method can straightforwardly model dependencies between any two tokens and force the model to pay attention to every token, which is more efficient for representational learning.
Under the longer training schedule (800 epochs), our model reaches 83.9\% accuracy, 0.4\% higher than MAE~\cite{he2021masked} and 0.9\% higher than RandSAC~\cite{hua2022self} (a concurrent autoregressive work of ours).

Though we focus on learning representations that
are better for fine-tuning, we also report the linear probing accuracy in \cref{tabimage_v2}.
For linear probing, we use the feature of the last block and adopt an extra BatchNorm~\cite{ioffe2015batch} layer before the linear classifier following MAE.

\begin{table*}[ht]
\begin{minipage}{\linewidth}
\centering
\caption{
\textbf{Comparison with previous on COCO and ADE20K.}
     $\ast$: the result is taken from~\cite{li2021benchmarking}, in which a grid-search was done to find the best hyperparameters for each method.
}
\label{tab:det}
\setlength{\tabcolsep}{7.2pt}
\scalebox{0.85}{
\begin{tabular}{llllll}
    \toprule
    \textbf{Method} & \textbf{Pretrain data} & \textbf{Pretrain epochs}& \textbf{COCO-AP$^{bb}$} & \textbf{COCO-AP$^{mk}$} & \textbf{ADE20K-mIOU}\\
    \midrule
    $\rm{supervised}^\ast$ & IN1K w/ labels &-& 47.9 & 42.9 &47.4 \\
    $\rm{MoCoV3}^\ast$~\cite{chen2021empirical} & IN1K &300& 47.9 & 42.7&47.3 \\
    $\rm{BEiT}^{\ast}$~\cite{bao2021beit} & IN1K+DALLE &1600& 49.8 & 44.4 &47.1 \\
    $\rm{MAE}^{\ast}$~\cite{he2021masked} & IN1K &1600&  50.3 &44.9&48.1  \\
    \midrule
    RandSAC~\cite{hua2022self} & IN1K &1600& - & - &47.3\\
    ViT-iGPT & IN1K & 300 &45.5	&41.2&	42.0 \\
    SAIM & IN1K &300& 47.4&42.8&	46.1\\
    SAIM & IN1K &800& 49.4 & 44.0 &47.8\\
    \bottomrule
\end{tabular}}
\end{minipage}
\hfill
\vspace{-10pt}
%
\end{table*}

\textbf{Object detection and instance segmentation on COCO}.
We conduct object detection and instance segmentation experiments on the MS COCO dataset ~\cite{lin2014microsoft}.
We adopt ViT~\cite{dosovitskiy2020image} as the backbone of Mask-RCNN~\cite{he2017mask}, following the architecture of ViT Benchmarking~\cite{li2021benchmarking}.
Implementation details are in Supplemental.
In ~\cref{tab:det}, we show the performance of representations learned through different self-supervised methods and supervised training.
We report box AP for object detection and mask AP for instance segmentation.
We observe that our method achieves $49.4$ bbox mAP and $44.0$ mask mAP.
The result is better than supervised learning and contrastive learning but lags behind MIM methods. We note that BEIT and MAE are pretrained with 1600 epochs and use grid-search to find the best hyperparameters, while we only pretrain 800 epochs and don't tune any parameters in the fine-tune stage due to limited access to computation.

\textbf{Semantic segmentation on ADE20K}.
We conduct semantic segmentation experiments on ADE20K dataset~\cite{zhou2019semantic}.
We adopt ViT~\cite{dosovitskiy2020image} as the backbone of UperNet~\cite{xiao2018unified}, following the implementation of BEiT~\cite{bao2021beit}.
Implementation details are in Supplemental.
In ~\cref{tab:det}, we show the performance of our method.
We report mIoU for semantic segmentation.
We observe that our method achieves 47.8 mIoU, which is slightly lower than MAE by 0.3, but higher than all others. 

In all tasks, our method surpasses the supervised results by large margins and achieves competitive performance with masked image modeling methods. Compared to the ViT-iGPT, our method improves 0.94\%, on Imagenet-1k,  4.07 mIOU on ADE20K and 1.9/1.6 mAP on COCO. All these results indicate that our method can learn high-quality representations, and  bridge the gap between MIM and AIM methods for unsupervised representation learning.

\subsection{Ablation study}
\label{sec:Ablation}
\textbf{Tokenization.} We first compare two different tokenization strategies introduced by iGPT~\cite{chen2020generative} and ViT~\cite{dosovitskiy2020image}, i.e., pixel-level modeling and patch-level modeling. ViT-style tokenization greatly improves the performance of the visual auto-aggressive method. The patch-based iGPT achieves 82.70 Top1 accuracy, which is higher than DeiT-base~\cite{touvron2021training}(81.8) but still lags behind the MIM method (e.g. 83.6 for MAE).

\begin{table*}[ht]
\centering
\caption{Ablation study on tokenization strategies, prediction order, decoder design, and Gaussian smoothing experiments by SAIM. $\ast$: the result is taken from ~\cite{chen2020generative}. }
\label{tab:ablation}
\begin{tabular}{cccccccc|c}
\toprule
\multicolumn{2}{c}{\textbf{Perform order}} & \multicolumn{2}{c}{\textbf{Decoder design}} & \multicolumn{2}{c}{\textbf{Gaussian smoothing}} & \multicolumn{2}{c}{\textbf{Tokenization}} & \multirow{2}{*}{\textbf{Fine-tune}} \\ \cline{1-8}
raster      & stochastic   & shared weight       & depth       & kernel size & sigma  &   pixel  & patch &            \\
\midrule
\multicolumn{5}{l}{\emph{Ablation of tokenization:}} \\ %
\Checkmark  &              & -   &      0          & -           & -      & \Checkmark  &  & 72.90$^{\ast}$     \\
\Checkmark  &              & -   &      0          & -           & -      & & \Checkmark &   \textbf{82.70}     \\

\midrule
\multicolumn{5}{l}{\emph{Ablation of prediction order:}} \\ %
\Checkmark  &              & -   &      0          & -           & -      & & \Checkmark  &   82.70     \\
\Checkmark&    &   \XSolidBrush  &  12    & -           & -      & & \Checkmark  &   82.76     \\
& \Checkmark   &   \XSolidBrush  &  12    & -           & -      & & \Checkmark  &   \textbf{83.40}     \\
\midrule
\multicolumn{3}{l}{\emph{Ablation of decoder design:}} \\ %
& \Checkmark   &   \XSolidBrush  &  12    & -           & -      & & \Checkmark  &   \textbf{83.40}     \\
& \Checkmark   & \Checkmark   &  12              & -           & -      & & \Checkmark  &   83.02     \\
            & \Checkmark   &       \XSolidBrush      &   6   & -           & -      & & \Checkmark   &   83.01       \\
            & \Checkmark   &    \XSolidBrush          &  1   & -           & -      & & \Checkmark   &   82.76         \\
\bottomrule
\multicolumn{3}{l}{\emph{Ablation of Gaussian smoothing:}} \\ %
& \Checkmark   &   \XSolidBrush  &  12    & -           & -      & & \Checkmark  &   83.40     \\
            & \Checkmark   &      \XSolidBrush        &   12   & 3           & 1      & & \Checkmark  &   83.52     \\
            & \Checkmark   &     \XSolidBrush         &  12   & 9           & 1      & & \Checkmark  &   \textbf{83.64}     \\
            & \Checkmark   &      \XSolidBrush        &  12    & 9           & 1.5    & & \Checkmark  &   83.45     \\
\midrule
\end{tabular}
\end{table*}

\textbf{Prediction order}. We compare the influence of different prediction orders. 
Our stochastic-order strategy (83.40) outperforms raster-order (82.70) 0.7\% in accuracy, which indicates that the stochastic-order strategy plays a key role in our method.
Our explanation here is that it leverages the richness and diversity of the visual signal and models full dependencies between all patches.

\textbf{Decoder design}.
We first try to verify the effect of applying parallel architecture, in which the decoder doesn't share weights with the encoder.
As is shown in the result, the parallel architecture gets a higher accuracy than the coupling architecture 0.4\%. 
By this design, the encoder can focus on learning semantic representations without participating in image generation.
We also vary the decoder depth to quantify its influence and find a sufficiently deep decoder performs better.

\textbf{Gaussian kernel parameters}.
We ablate Gaussian kernel parameters in~\cref{tab:ablation}.
We investigate the kernel size and standard deviation in Gaussian distribution.
The larger the convolution kernel size or standard deviation is, the more texture will be removed from the target image.
We obtain the best result when the kernel size is 9 and the standard deviation is 1, and all experiments with Gaussian smooth outperform those without.
\section{Visualization and analysis}
\label{sec:Visualization}

\begin{figure*}[ht]
    \centering
    \includegraphics[width=1\textwidth]{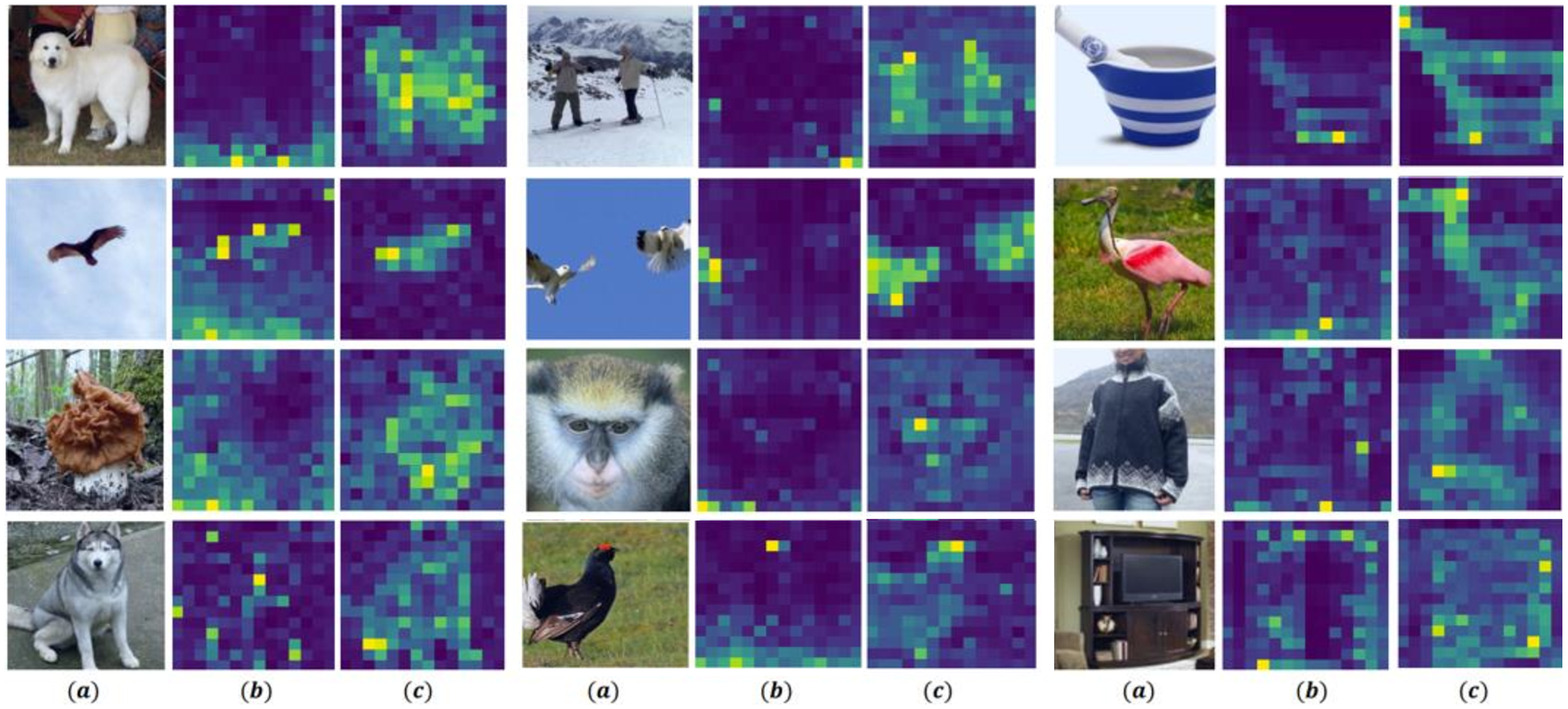}
    \caption{
        \textbf{Attention maps of different prediction order.}
        We show example results for ImageNet validation set.
        Description of images from left to right: (a) the original image, (b) the attention map of raster-order AIM, (c) the attention map of SAIM, more examples are in the appendix.
    }
    \label{fig:attention}
\end{figure*}

\begin{figure*}[!h]
    \centering
    \includegraphics[width=1\textwidth]{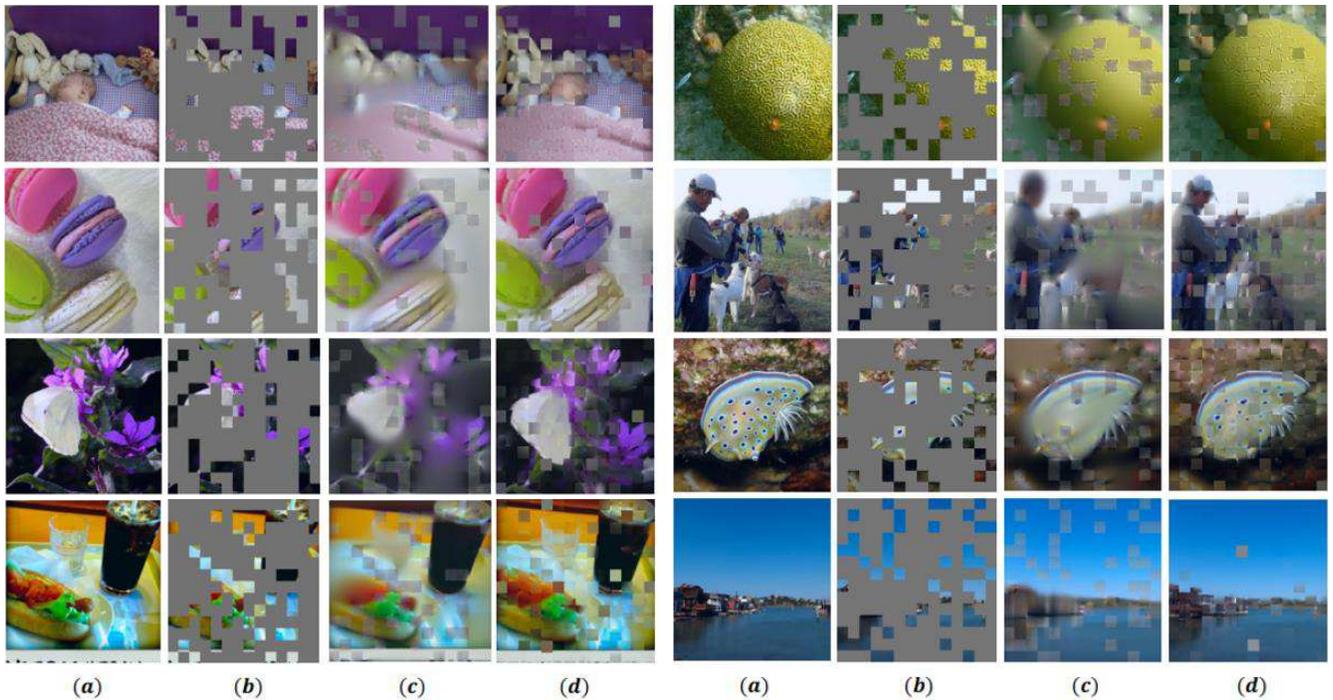}
    \caption{
        \textbf{Reconstruction image of different methods.}
        We show example results for ImageNet validation set. Description of images from left to right: (a) the origin image, (b) the mask map of the MAE, (c) the reconstruction image of MAE, (d) the reconstruction image of SAIM, more examples are in the appendix.
    }
    \label{fig:reconstruct}
\end{figure*}

\textbf{Autoregressive image modeling attention maps.} 
As shown in~\cref{fig:attention}, we observed that the fixed-order autoregressive model cannot pay attention to the main object of the input image, and is only concerned with the local area of the image.
While, SAIM with stochastic order focuses on the main information of the image, and obtains human-level attention representation with unlabeled data.
In the field of computer vision, images are high-dimensional spatial information, and the main information of the input image is randomly distributed in tokens.
SAIM allows each prediction token can have the opportunity to see the global context information, which is crucial for autoregressive image modeling.

\textbf{The reconstruction of autoregressive and autoencoder.}
As shown in~\cref{fig:reconstruct}, we discovered that MAE masked out 75\% of the image tokens, resulting in blurred reconstruction results in large masked regions.
According to the MAE, reducing the number of image masks can improve reconstruction results, but it will decrease model representation due to image local dependency.
While, our proposed stochastic autoregressive image modeling, SAIM, utilizes all the information of the image to generate clear images, and achieve better fine-tuning accuracy than MAE on ImageNet-1K.

For example, the first quadruple in ~\cref{fig:reconstruct}, shows that the baby in the image is ignored by MAE because it's totally masked, but SAIM can efficiently generate the baby in the image.
Therefore, if the subject feature of the image is totally masked, the model will misunderstand the subject of the image.
This problem will be harmful to downstream tasks.
We argue that autoencoder methods introduce independence assumptions in image tasks, e.g., the masked subset of patches can't establish decencies with each other.
The stochastic autoregressive image model enables the model to obtain the global dependency of the image by establishing the global context of the image.

\section{Conclusion}\label{sec:conclusion}
In this paper, we fully analyze the differences between autoregressive modeling on visual and textual tasks and propose how to improve the performance of visual autoregressive models.
Therefore, we explore a novel stochastic autoregressive image modeling (SAIM).
By introducing ViT-style tokenization, stochastic order prediction, and the parallel encoder-decoder, we significantly improve the performance of autoregressive image modeling and bridge the gap between computer vision and NLP.
Experiments on downstream tasks(image classification, object detection, and semantic segmentation) demonstrate the effectiveness of our method.
%
Our method achieves competitive performance compared with other self-supervised methods, this proved that the autoregressive image modeling is also an effective pretext task for visual representation learning. 

\textbf{Discussion.} The major drawback of this work is that our approach is time-consuming since our parallel architecture needs to calculate the self-attention blocks twice. Designing a lightweight decoder is challenging but worth exploring. We will conduct future studies on this issue.

{
\small

\bibliography{ref}

\begin{thebibliography}{45}
\providecommand{\natexlab}[1]{#1}

\bibitem[{Baevski et~al.(2022)Baevski, Hsu, Xu, Babu, Gu, and
  Auli}]{baevski2022data2vec}
Baevski, A.; Hsu, W.-N.; Xu, Q.; Babu, A.; Gu, J.; and Auli, M. 2022.
\newblock Data2vec: A general framework for self-supervised learning in speech,
  vision and language.
\newblock \emph{arXiv preprint arXiv:2202.03555}.

\bibitem[{Bao, Dong, and Wei(2021)}]{bao2021beit}
Bao, H.; Dong, L.; and Wei, F. 2021.
\newblock Beit: Bert pre-training of image transformers.
\newblock \emph{arXiv preprint arXiv:2106.08254}.

\bibitem[{Brown et~al.(2020)Brown, Mann, Ryder, Subbiah, Kaplan, Dhariwal,
  Neelakantan, Shyam, Sastry, Askell et~al.}]{brown2020language}
Brown, T.; Mann, B.; Ryder, N.; Subbiah, M.; Kaplan, J.~D.; Dhariwal, P.;
  Neelakantan, A.; Shyam, P.; Sastry, G.; Askell, A.; et~al. 2020.
\newblock Language models are few-shot learners.
\newblock \emph{Advances in neural information processing systems}, 33:
  1877--1901.

\bibitem[{Caron et~al.(2021)Caron, Touvron, Misra, J{\'e}gou, Mairal,
  Bojanowski, and Joulin}]{caron2021emerging}
Caron, M.; Touvron, H.; Misra, I.; J{\'e}gou, H.; Mairal, J.; Bojanowski, P.;
  and Joulin, A. 2021.
\newblock Emerging properties in self-supervised vision transformers.
\newblock In \emph{Proceedings of the IEEE/CVF International Conference on
  Computer Vision}, 9650--9660.

\bibitem[{Chen et~al.(2020{\natexlab{a}})Chen, Radford, Child, Wu, Jun, Luan,
  and Sutskever}]{chen2020generative}
Chen, M.; Radford, A.; Child, R.; Wu, J.; Jun, H.; Luan, D.; and Sutskever, I.
  2020{\natexlab{a}}.
\newblock Generative pretraining from pixels.
\newblock In \emph{International Conference on Machine Learning}, 1691--1703.
  PMLR.

\bibitem[{Chen et~al.(2020{\natexlab{b}})Chen, Kornblith, Norouzi, and
  Hinton}]{chen2020simple}
Chen, T.; Kornblith, S.; Norouzi, M.; and Hinton, G. 2020{\natexlab{b}}.
\newblock A simple framework for contrastive learning of visual
  representations.
\newblock In \emph{International conference on machine learning}, 1597--1607.
  PMLR.

\bibitem[{Chen et~al.(2020{\natexlab{c}})Chen, Kornblith, Swersky, Norouzi, and
  Hinton}]{chen2020big}
Chen, T.; Kornblith, S.; Swersky, K.; Norouzi, M.; and Hinton, G.~E.
  2020{\natexlab{c}}.
\newblock Big self-supervised models are strong semi-supervised learners.
\newblock \emph{Advances in neural information processing systems}, 33:
  22243--22255.

\bibitem[{Chen et~al.(2021)Chen, Saxena, Li, Fleet, and
  Hinton}]{chen2021pix2seq}
Chen, T.; Saxena, S.; Li, L.; Fleet, D.~J.; and Hinton, G. 2021.
\newblock Pix2seq: A language modeling framework for object detection.
\newblock \emph{arXiv preprint arXiv:2109.10852}.

\bibitem[{Chen et~al.(2022)Chen, Ding, Wang, Xin, Mo, Wang, Han, Luo, Zeng, and
  Wang}]{chen2022context}
Chen, X.; Ding, M.; Wang, X.; Xin, Y.; Mo, S.; Wang, Y.; Han, S.; Luo, P.;
  Zeng, G.; and Wang, J. 2022.
\newblock Context autoencoder for self-supervised representation learning.
\newblock \emph{arXiv preprint arXiv:2202.03026}.

\bibitem[{Chen, Xie, and He(2021)}]{chen2021empirical}
Chen, X.; Xie, S.; and He, K. 2021.
\newblock An empirical study of training self-supervised visual transformers.
\newblock \emph{arXiv e-prints}, arXiv--2104.

\bibitem[{Devlin et~al.(2018)Devlin, Chang, Lee, and
  Toutanova}]{devlin2018bert}
Devlin, J.; Chang, M.-W.; Lee, K.; and Toutanova, K. 2018.
\newblock Bert: Pre-training of deep bidirectional transformers for language
  understanding.
\newblock \emph{arXiv preprint arXiv:1810.04805}.

\bibitem[{Dosovitskiy et~al.(2020)Dosovitskiy, Beyer, Kolesnikov, Weissenborn,
  Zhai, Unterthiner, Dehghani, Minderer, Heigold, Gelly
  et~al.}]{dosovitskiy2020image}
Dosovitskiy, A.; Beyer, L.; Kolesnikov, A.; Weissenborn, D.; Zhai, X.;
  Unterthiner, T.; Dehghani, M.; Minderer, M.; Heigold, G.; Gelly, S.; et~al.
  2020.
\newblock An image is worth 16x16 words: Transformers for image recognition at
  scale.
\newblock \emph{arXiv preprint arXiv:2010.11929}.

\bibitem[{Gidaris, Singh, and Komodakis(2018)}]{gidaris2018unsupervised}
Gidaris, S.; Singh, P.; and Komodakis, N. 2018.
\newblock Unsupervised representation learning by predicting image rotations.
\newblock \emph{arXiv preprint arXiv:1803.07728}.

\bibitem[{Goyal et~al.(2017)Goyal, Doll{\'a}r, Girshick, Noordhuis, Wesolowski,
  Kyrola, Tulloch, Jia, and He}]{goyal2017accurate}
Goyal, P.; Doll{\'a}r, P.; Girshick, R.; Noordhuis, P.; Wesolowski, L.; Kyrola,
  A.; Tulloch, A.; Jia, Y.; and He, K. 2017.
\newblock Accurate, large minibatch sgd: Training imagenet in 1 hour.
\newblock \emph{arXiv preprint arXiv:1706.02677}.

\bibitem[{Grill et~al.(2020)Grill, Strub, Altch{\'e}, Tallec, Richemond,
  Buchatskaya, Doersch, Avila~Pires, Guo, Gheshlaghi~Azar
  et~al.}]{grill2020bootstrap}
Grill, J.-B.; Strub, F.; Altch{\'e}, F.; Tallec, C.; Richemond, P.;
  Buchatskaya, E.; Doersch, C.; Avila~Pires, B.; Guo, Z.; Gheshlaghi~Azar, M.;
  et~al. 2020.
\newblock Bootstrap your own latent-a new approach to self-supervised learning.
\newblock \emph{Advances in Neural Information Processing Systems}, 33:
  21271--21284.

\bibitem[{He et~al.(2021)He, Chen, Xie, Li, Doll{\'a}r, and
  Girshick}]{he2021masked}
He, K.; Chen, X.; Xie, S.; Li, Y.; Doll{\'a}r, P.; and Girshick, R. 2021.
\newblock Masked autoencoders are scalable vision learners.
\newblock \emph{arXiv preprint arXiv:2111.06377}.

\bibitem[{He et~al.(2020)He, Fan, Wu, Xie, and Girshick}]{he2020momentum}
He, K.; Fan, H.; Wu, Y.; Xie, S.; and Girshick, R. 2020.
\newblock Momentum contrast for unsupervised visual representation learning.
\newblock In \emph{Proceedings of the IEEE/CVF conference on computer vision
  and pattern recognition}, 9729--9738.

\bibitem[{He et~al.(2017)He, Gkioxari, Doll{\'a}r, and Girshick}]{he2017mask}
He, K.; Gkioxari, G.; Doll{\'a}r, P.; and Girshick, R. 2017.
\newblock Mask r-cnn.
\newblock In \emph{Proceedings of the IEEE international conference on computer
  vision}, 2961--2969.

\bibitem[{Hua et~al.(2022)Hua, Tian, Ren, Zhao, and Sigal}]{hua2022self}
Hua, T.; Tian, Y.; Ren, S.; Zhao, H.; and Sigal, L. 2022.
\newblock Self-supervision through Random Segments with Autoregressive Coding
  (RandSAC).
\newblock \emph{arXiv preprint arXiv:2203.12054}.

\bibitem[{Ioffe and Szegedy(2015)}]{ioffe2015batch}
Ioffe, S.; and Szegedy, C. 2015.
\newblock Batch normalization: Accelerating deep network training by reducing
  internal covariate shift.
\newblock In \emph{International conference on machine learning}, 448--456.
  PMLR.

\bibitem[{Koch et~al.(2015)Koch, Zemel, Salakhutdinov et~al.}]{koch2015siamese}
Koch, G.; Zemel, R.; Salakhutdinov, R.; et~al. 2015.
\newblock Siamese neural networks for one-shot image recognition.
\newblock In \emph{ICML deep learning workshop}, volume~2, 0. Lille.

\bibitem[{Li et~al.(2021)Li, Xie, Chen, Dollar, He, and
  Girshick}]{li2021benchmarking}
Li, Y.; Xie, S.; Chen, X.; Dollar, P.; He, K.; and Girshick, R. 2021.
\newblock Benchmarking detection transfer learning with vision transformers.
\newblock \emph{arXiv preprint arXiv:2111.11429}.

\bibitem[{Lin et~al.(2014)Lin, Maire, Belongie, Hays, Perona, Ramanan,
  Doll{\'a}r, and Zitnick}]{lin2014microsoft}
Lin, T.-Y.; Maire, M.; Belongie, S.; Hays, J.; Perona, P.; Ramanan, D.;
  Doll{\'a}r, P.; and Zitnick, C.~L. 2014.
\newblock Microsoft coco: Common objects in context.
\newblock In \emph{European conference on computer vision}, 740--755. Springer.

\bibitem[{Loshchilov and Hutter(2016)}]{loshchilov2016sgdr}
Loshchilov, I.; and Hutter, F. 2016.
\newblock Sgdr: Stochastic gradient descent with warm restarts.
\newblock \emph{arXiv preprint arXiv:1608.03983}.

\bibitem[{Loshchilov and Hutter(2017)}]{loshchilov2017decoupled}
Loshchilov, I.; and Hutter, F. 2017.
\newblock Decoupled weight decay regularization.
\newblock \emph{arXiv preprint arXiv:1711.05101}.

\bibitem[{Noroozi and Favaro(2016)}]{noroozi2016unsupervised}
Noroozi, M.; and Favaro, P. 2016.
\newblock Unsupervised learning of visual representations by solving jigsaw
  puzzles.
\newblock In \emph{European conference on computer vision}, 69--84. Springer.

\bibitem[{Oord, Li, and Vinyals(2018)}]{oord2018representation}
Oord, A. v.~d.; Li, Y.; and Vinyals, O. 2018.
\newblock Representation learning with contrastive predictive coding.
\newblock \emph{arXiv preprint arXiv:1807.03748}.

\bibitem[{Pathak et~al.(2016)Pathak, Krahenbuhl, Donahue, Darrell, and
  Efros}]{pathak2016context}
Pathak, D.; Krahenbuhl, P.; Donahue, J.; Darrell, T.; and Efros, A.~A. 2016.
\newblock Context encoders: Feature learning by inpainting.
\newblock In \emph{Proceedings of the IEEE conference on computer vision and
  pattern recognition}, 2536--2544.

\bibitem[{Radford et~al.(2018)Radford, Narasimhan, Salimans, and
  Sutskever}]{radford2018improving}
Radford, A.; Narasimhan, K.; Salimans, T.; and Sutskever, I. 2018.
\newblock Improving language understanding by generative pre-training.

\bibitem[{Radford et~al.(2019)Radford, Wu, Child, Luan, Amodei, Sutskever
  et~al.}]{radford2019language}
Radford, A.; Wu, J.; Child, R.; Luan, D.; Amodei, D.; Sutskever, I.; et~al.
  2019.
\newblock Language models are unsupervised multitask learners.
\newblock \emph{OpenAI blog}, 1(8): 9.

\bibitem[{Ramesh et~al.(2021)Ramesh, Pavlov, Goh, Gray, Voss, Radford, Chen,
  and Sutskever}]{ramesh2021zero}
Ramesh, A.; Pavlov, M.; Goh, G.; Gray, S.; Voss, C.; Radford, A.; Chen, M.; and
  Sutskever, I. 2021.
\newblock Zero-shot text-to-image generation.
\newblock In \emph{International Conference on Machine Learning}, 8821--8831.
  PMLR.

\bibitem[{Russakovsky et~al.(2015)Russakovsky, Deng, Su, Krause, Satheesh, Ma,
  Huang, Karpathy, Khosla, Bernstein et~al.}]{russakovsky2015imagenet}
Russakovsky, O.; Deng, J.; Su, H.; Krause, J.; Satheesh, S.; Ma, S.; Huang, Z.;
  Karpathy, A.; Khosla, A.; Bernstein, M.; et~al. 2015.
\newblock Imagenet large scale visual recognition challenge.
\newblock \emph{International journal of computer vision}, 115(3): 211--252.

\bibitem[{Szegedy et~al.(2016)Szegedy, Vanhoucke, Ioffe, Shlens, and
  Wojna}]{szegedy2016rethinking}
Szegedy, C.; Vanhoucke, V.; Ioffe, S.; Shlens, J.; and Wojna, Z. 2016.
\newblock Rethinking the inception architecture for computer vision.
\newblock In \emph{Proceedings of the IEEE conference on computer vision and
  pattern recognition}, 2818--2826.

\bibitem[{Touvron et~al.(2021)Touvron, Cord, Douze, Massa, Sablayrolles, and
  J{\'e}gou}]{touvron2021training}
Touvron, H.; Cord, M.; Douze, M.; Massa, F.; Sablayrolles, A.; and J{\'e}gou,
  H. 2021.
\newblock Training data-efficient image transformers \& distillation through
  attention.
\newblock In \emph{International Conference on Machine Learning}, 10347--10357.
  PMLR.

\bibitem[{Van~den Oord et~al.(2016)Van~den Oord, Kalchbrenner, Espeholt,
  Vinyals, Graves et~al.}]{van2016conditional}
Van~den Oord, A.; Kalchbrenner, N.; Espeholt, L.; Vinyals, O.; Graves, A.;
  et~al. 2016.
\newblock Conditional image generation with pixelcnn decoders.
\newblock \emph{Advances in neural information processing systems}, 29.

\bibitem[{Van Den~Oord, Vinyals et~al.(2017)}]{van2017neural}
Van Den~Oord, A.; Vinyals, O.; et~al. 2017.
\newblock Neural discrete representation learning.
\newblock \emph{Advances in neural information processing systems}, 30.

\bibitem[{Van~Oord, Kalchbrenner, and Kavukcuoglu(2016)}]{van2016pixel}
Van~Oord, A.; Kalchbrenner, N.; and Kavukcuoglu, K. 2016.
\newblock Pixel recurrent neural networks.
\newblock In \emph{International conference on machine learning}, 1747--1756.
  PMLR.

\bibitem[{Vaswani et~al.(2017)Vaswani, Shazeer, Parmar, Uszkoreit, Jones,
  Gomez, Kaiser, and Polosukhin}]{vaswani2017attention}
Vaswani, A.; Shazeer, N.; Parmar, N.; Uszkoreit, J.; Jones, L.; Gomez, A.~N.;
  Kaiser, {\L}.; and Polosukhin, I. 2017.
\newblock Attention is all you need.
\newblock \emph{Advances in neural information processing systems}, 30.

\bibitem[{Wei et~al.(2021)Wei, Fan, Xie, Wu, Yuille, and
  Feichtenhofer}]{wei2021masked}
Wei, C.; Fan, H.; Xie, S.; Wu, C.-Y.; Yuille, A.; and Feichtenhofer, C. 2021.
\newblock Masked Feature Prediction for Self-Supervised Visual Pre-Training.
\newblock \emph{arXiv preprint arXiv:2112.09133}.

\bibitem[{Xiao et~al.(2018)Xiao, Liu, Zhou, Jiang, and Sun}]{xiao2018unified}
Xiao, T.; Liu, Y.; Zhou, B.; Jiang, Y.; and Sun, J. 2018.
\newblock Unified perceptual parsing for scene understanding.
\newblock In \emph{Proceedings of the European Conference on Computer Vision
  (ECCV)}, 418--434.

\bibitem[{Xie et~al.(2021)Xie, Zhang, Cao, Lin, Bao, Yao, Dai, and
  Hu}]{xie2021simmim}
Xie, Z.; Zhang, Z.; Cao, Y.; Lin, Y.; Bao, J.; Yao, Z.; Dai, Q.; and Hu, H.
  2021.
\newblock Simmim: A simple framework for masked image modeling.
\newblock \emph{arXiv preprint arXiv:2111.09886}.

\bibitem[{Yang et~al.(2019)Yang, Dai, Yang, Carbonell, Salakhutdinov, and
  Le}]{yang2019xlnet}
Yang, Z.; Dai, Z.; Yang, Y.; Carbonell, J.; Salakhutdinov, R.~R.; and Le, Q.~V.
  2019.
\newblock Xlnet: Generalized autoregressive pretraining for language
  understanding.
\newblock \emph{Advances in neural information processing systems}, 32.

\bibitem[{Zhang, Isola, and Efros(2016)}]{zhang2016colorful}
Zhang, R.; Isola, P.; and Efros, A.~A. 2016.
\newblock Colorful image colorization.
\newblock In \emph{European conference on computer vision}, 649--666. Springer.

\bibitem[{Zhou et~al.(2019)Zhou, Zhao, Puig, Xiao, Fidler, Barriuso, and
  Torralba}]{zhou2019semantic}
Zhou, B.; Zhao, H.; Puig, X.; Xiao, T.; Fidler, S.; Barriuso, A.; and Torralba,
  A. 2019.
\newblock Semantic understanding of scenes through the ade20k dataset.
\newblock \emph{International Journal of Computer Vision}, 127(3): 302--321.

\bibitem[{Zhou et~al.(2021)Zhou, Wei, Wang, Shen, Xie, Yuille, and
  Kong}]{zhou2021ibot}
Zhou, J.; Wei, C.; Wang, H.; Shen, W.; Xie, C.; Yuille, A.; and Kong, T. 2021.
\newblock ibot: Image bert pre-training with online tokenizer.
\newblock \emph{arXiv preprint arXiv:2111.07832}.

\end{thebibliography}
}
\appendix

\section{Additional Ablations}\label{sec:Additional_Ablations}
We discussed the implications of prediction order, decoupling effect, Gaussian kernel parameters, and decoder depth on SAIM in the main paper.
The types of loss functions and the project head on SAIM are shown below.

\textbf{Loss function}. The encoder in our SAIM maps the image to the latent representation which is then used by the decoder to generate the original image.
We calculate the pixel similarity between the output and input images with gaussian processing.
As shown in the ~\cref{tab:ablation}, we used several types of loss functions, including L1, mean squared error(MSE) and mean squared error(MSE) with the normalized pixel.
The MSE with the normalized pixel obtains the best results in the experiments.

\textbf{Predcition head}. The prediction head uses the latent representation to generate the input image.
The output channels of the projection head equal the number of pixels in the patch.
The types of the prediction head have different influences on SAIM.
In this work, we conduct three types separately: a MLP layer, a linear layer, and two transformer layers ~\cite{vaswani2017attention}.
As shown in the ~\cref{tab:ablation},  SAIM with one layer MLP obtains the best result in this experiment.
%

\begin{table*}[h!]
\centering
\caption{Ablation study on the types of loss functions and projection head with 300 epochs on SAIM. $\dag$:is MSE with the normalized pixel.}
\label{tab:ablation}
\begin{tabular}{ccccccc|c}
\toprule
\multicolumn{3}{c}{\textbf{Loss Type}} & \multicolumn{3}{c}{\textbf{Projection Head}} & & \multirow{2}{*}{\textbf{Fine-tune}} \\ \cline{1-6}  
L1      & MSE   & $\rm{MSE^{\dag}}$     & Linear       & MLP & Transformer  &   & \\
\midrule
\multicolumn{5}{l}{\emph{Ablation of loss function:}} \\ %
\checkmark  &              &              &   & \checkmark  &  &   &   83.02     \\
            & \checkmark   &              &   & \checkmark  &  &   &   83.14     \\
            &              &  \checkmark  &   & \checkmark  &  &   &   83.64     \\
\midrule
\multicolumn{3}{l}{\emph{Ablation of prediction head:}} \\ %
            &    &  \checkmark  & \checkmark     &              &             &     &   83.55     \\
            &    &  \checkmark  &                & \checkmark   &             &     &   83.64     \\
            &    &  \checkmark  &                &              & \checkmark  &     &   83.50     \\
\bottomrule
\end{tabular}
\end{table*}

\section{More setup}\label{sec:More_setup}

\textbf{Longer training.} As shown in the ~\cref{longer}, we observe that longer training enhances the performance of SAIM with the backbone on ViT-Base ~\cite{dosovitskiy2020image}.
With an increase from 300 epochs to 800/1600 epochs, the results of downstream tasks will be improved.
We believe that better experimental results can be achieved with a longer training time.
\begin{table*}[h]
\centering
    \caption{ \textbf{Impact of longer training.} }
\small
  \setlength{\tabcolsep}{16pt}
    \begin{tabular}{@{} l c c  c c c}
      \toprule
	    Method     & Architecture  & epochs  & ImageNet-1K & COCO aa& ADE20K\\

        \midrule
	    {SAIM} & {ViT-Base} & 300  &     83.6&42.8&	46.1 \\
         {SAIM}&{ViT-Base}  & 800  &  83.9&44.0&	47.8\\
          {SAIM}&  {ViT-Base}& 1600  &  83.9&44.5&	48.0
\\
      \bottomrule
 \end{tabular}
  \label{longer}
\end{table*}

\textbf{Computation resources.} Our SAIM is trained on 32-V100 GPUs.
Training for 300 epochs needs a day and a half and 4 days for 800 epochs.
In downstream experiments, the object detection task uses 32-V100 GPUs, which takes 3 days for 100 epochs.
The semantic segmentation task uses 8-V100 GPUs, which takes one day for 160k iterations.
We set different random seeds to report the average results after multiple experiments.

\section{Hyperparameters}\label{sec:Hyperparameters}
\subsection{Hyperparameters for pretraing on ImageNet-1K}

\textbf{Pre-training on ImageNet-1K}. The hyperparameters used for pretraining SAIM are shown in the ~\cref{tab:pretrain_hyperparameters}.
We followed the pre-training parameters in MAE~\cite{he2021masked} and MaskFeat~\cite{wei2021masked} without using color jittering, drop path, and gradient clip.
We employ an AdamW~\cite{loshchilov2017decoupled} optimizer with a cosine learning rate scheduler~\cite{loshchilov2016sgdr}.
The batch size is 2048, the warmup epoch~\cite{goyal2017accurate} is 30 and the weight decay is 0.05.

\begin{table}[h]
\centering
\caption{\textbf{Hyperparameters for pretraining.}}
\label{tab:pretrain_hyperparameters}
\begin{tabular}{l|l}
\cline{1-2}
config                 & value                    \\ \cline{1-2}
optimizer              & AdamW                    \\
base learning rate     & 2e-4                   \\
weight decay           & 0.05                     \\
optimizer momentum     & $\beta_{1}$,  $\beta_{2}$=0.9, 0.95 ~\cite{chen2020generative}  \\
batch size             & 2048                     \\
learning rate schedule & cosine decay             \\
warmup epochs          & 30                       \\
augmentation           & RandomResizeCrop         \\ \cline{1-2}
\end{tabular}
\end{table}

\subsection{Hyperparameters for finetuning on ImageNet-1k.}
\textbf{Fine-tuning on ImageNet-1k}. The hyperparameters used for finetuning SAIM on ImageNet-1K classification are shown in the ~\cref{tab:finetune_hyperparameters}. 
We use layer-wise learning rate decay, weight decay, and AdamW.
The batch size is 1024, the warmup epoch is 20 and the weight decay is 0.05. For ViT-Base, we train 100 epochs with a learning rate of 5e-4 and a layer-wise decay rate of 0.65.

\begin{table}[h]
\centering
\caption{\textbf{Hyperparameters for finetuning on ImageNet-1k.}}
\label{tab:finetune_hyperparameters}
\begin{tabular}{l|l}
\hline
config                 & value                           \\ \hline
optimizer              & AdamW                           \\
base learning rate     & 5e-4                            \\
weight decay           & 0.05                            \\
optimizer momentum     & $\beta_1$,  $\beta_2$ = 0.9, 0.99 \\
layer-wise lr decay    & 0.65                            \\
batch size             & 1024                            \\
learning rate schedule & cosine decay                    \\
warmup epochs          & 20                               \\
training epoch         & 100                             \\
augmentation           & RandAug       (9, 0.5)                 \\
label smoothing~\cite{szegedy2016rethinking}        & 0.1                             \\
mixup                  & 0.8                             \\
cutmix                 & 1.0                             \\
drop path              & 0.1                             \\ \hline
\end{tabular}
\end{table}

\subsection{Hyperparameters for object detection and instance segmentation on COCO}
\textbf{Object detection and instance segmentation on COCO}. The hyperparameters used for finetuning SAIM on COCO~\cite{lin2014microsoft} are shown in the ~\cref{tab:object detection and instance segmentation on COCO}. 
The batch size is 64, the warmup epoch is 0.25, and the weight decay is 0.1. For ViT-Base, we train 100 epochs with a learning rate of 8e-5.

\begin{table}[ht]
\centering
\caption{\textbf{Hyperparameters for object detection and instance segmentation on COCO.}}
\label{tab:object detection and instance segmentation on COCO}
\begin{tabular}{l|l}
\hline
config                 & value                           \\ \hline
optimizer              & AdamW                           \\
learning rate          & 8e-5                            \\
weight decay           & 0.1                           \\
optimizer momentum     & $\beta_1$,  $\beta_2$ = 0.9, 0.99 \\
batch size             & 64                            \\
learning rate schedule & cosine decay                    \\
warmup epochs          & 0.25                               \\
training epoch         & 100                             \\
drop path              & 0.1                             \\ 
input resolution       & 1024 × 1024    \\
position embedding interpolate  & bilinear \\
\hline
\end{tabular}
\end{table}

\subsection{Hyperparameters for semantic segmentation on ADE20K}
\textbf{semantic segmentation on ADE20K}. The hyperparameters used for finetuning SAIM on ADE20K~\cite{zhou2019semantic} are shown in the ~\cref{tab:semantic segmentation on ADE20K}. 
We use layer-wise learning rate decay, weight decay, and AdamW.
The batch size is 16, the warmup iteration is 1500 and the weight decay is 0.05. For ViT-Base, we train 160k iterations with a learning rate of 4e-4.

\begin{table}[ht]
\centering
\caption{\textbf{Hyperparameters for semantic segmentation on ADE20K.}}
\label{tab:semantic segmentation on ADE20K}
\begin{tabular}{l|l}
\hline
config                 & value                           \\ \hline
optimizer              & AdamW                           \\
learning rate          & 4e-4                            \\
weight decay           & 0.05                        \\
layer-wise lr decay    & 0.65                            \\
optimizer momentum     & $\beta_1$,  $\beta_2$ = 0.9, 0.99 \\
batch size             & 16                            \\
learning rate schedule & poly decay                    \\
warmup iterations          & 1500                               \\
training iterations         & 160k                           \\
drop path              & 0.1                             \\ 
input resolution       & 512 × 512    \\
position embedding interpolate  & bilinear \\
\hline
\end{tabular}
\end{table}

\section{Pseudo-code}\label{sec:Pseudo_code}
To make our SAIM easy to understand, we provide pseudo-code in a Pytorch-like style. To simplify, the SAIM is implemented in ~\cref{alg:code}.

\begin{algorithm*}[h]
\caption{Pseudocode of SAIM in a PyTorch-like style.}
\label{alg:code}
\definecolor{codeblue}{rgb}{0.25,0.5,0.5}
\lstset{
  backgroundcolor=\color{white},
  basicstyle=\fontsize{7.2pt}{7.2pt}\ttfamily\selectfont,
  columns=fullflexible,
  breaklines=true,
  captionpos=b,
  commentstyle=\fontsize{7.2pt}{7.2pt}\color{codeblue},
  keywordstyle=\fontsize{7.2pt}{7.2pt},
}
\begin{lstlisting}[language=python]
# imgs: a minibatch with N samples
# pos_embed: fixed 2d sin_cos pos_embed
# P: patch size

# depth: total number of encoder(decoder) blocks
# encoder_blocks: "depth" layers of self-attention blocks
# decoder_blocks: "depth" layers of cross-attention blocks
# projector: a layer of MLP

# initialize g and h
N, C, W, H = imgs.shape
x = PatchEmbed(imgs)
h = x + pos_embed
g = pos_embed

# generate masks
N, L, D = h.shape
noise = torch.rand(N, L, device=x.device)  # uniform distribution in [0, 1]
mask_h = noise.unsqueeze(-1) >= noise.unsqueeze(1) # broadcast to N*L*L
mask_g = noise.unsqueeze(-1) > noise.unsqueeze(1) # True represents visible

# forward
for i in range(depth):
    h = self.encoder_blocks[i](q=h, kv=h, mask=mask_h) # self-attention layer
    g = self.decoder_blocks[i](q=g, kv=h, mask=mask_g) # cross-attention layer
g = projector(g) # MLP layer

# compute loss
target = imgs.reshape(N,C,H//P,P,W//P,P).permute(0,2,4,3,5,1).reshape(N,-1,P*P*C)
loss = MSE(g, target)

# backward
loss.backward()
update(encoder_blocks, decoder_blocks, projector)
\end{lstlisting}
\end{algorithm*}

\section{Self-Attention Visualizations} \label{sec:Self-Attention}

According to several reference points, we use the attention map to visualize the last layer of the SAIM as shown in the ~\cref{fig:attention}.
The visualizations are produced by attention scores computed via query-key product in the last layer.
For each reference point, we use the corresponding patch as the query.
The attention map can distinguish objects without supervision. 
Therefore, this visualization indicates that SAIM learns useful knowledge of images.
The property partially indicates the reason why SAIM can help downstream tasks.

\begin{figure*}[ht]
    \centering
    \includegraphics[width=1\textwidth]{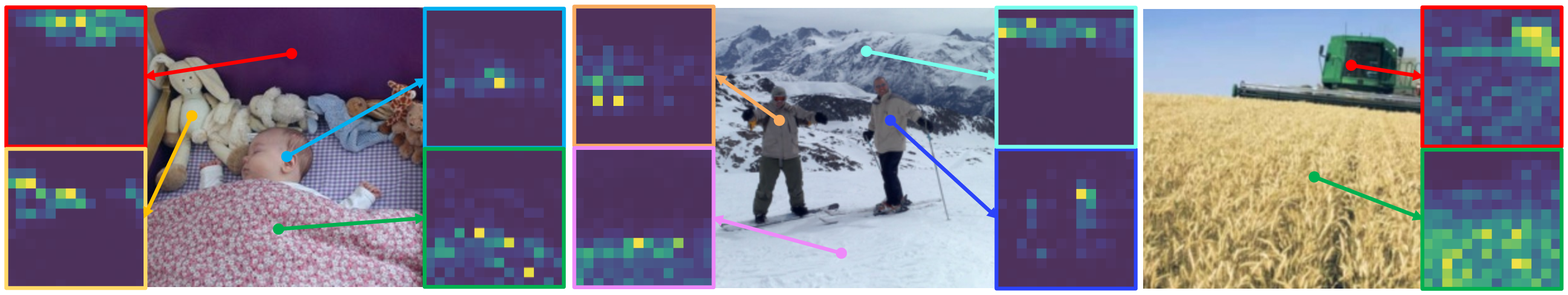}
    \caption{
        \textbf{Self-attention for a set of reference points.}
        We visualize the self-attention map from the last layer of a ViT-Base trained with SAIM.
        The network is able to separate objects, though it has been trained with no supervision at all.
    }
    \label{fig:attention}
\end{figure*}



\section{Visualizing Stochastic Order Attention Mask}\label{sec:Attention_mask}

\begin{figure*}[h]
    \centering
    \includegraphics[width=1\textwidth]{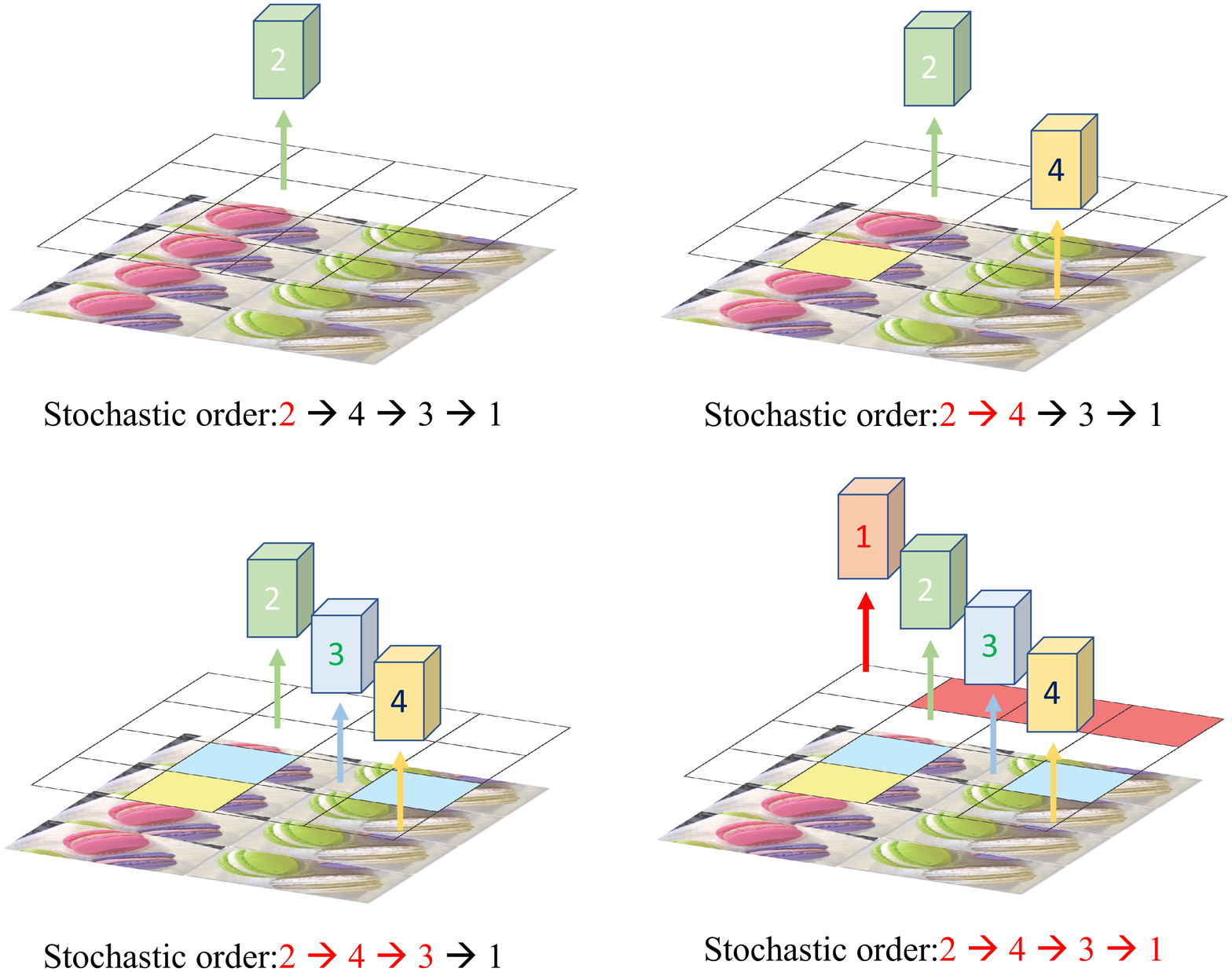}
    \caption{
        \textbf{Stochastic order of attention mask.}
        Illustration of the stochastic autoregressive image modeling.
    }
    \label{fig:attention_mask}
\end{figure*}

We provide a detailed visualization of the proposed stochastic order attention mask, including stochastic order, query attention mask, and how we use stochastic order attention mask to predict the current token in sequence. 
As shown in the ~\cref{fig:attention_mask}, SAIM predicts the current token according to the context of the visible token in stochastic order.
For example, the stochastic order is 2-> 4-> 3->1.
When we want to predict the third token, SAIM can see the second and fourth tokens, and so on.
Our stochastic strategy constructs an irregular attention mask that increases the richness and variety of the visible signals.

\section{Compare AIM and MIM from the perspective of conditional probability}\label{sec:Attention_mask}
Give an image $\boldsymbol x$, we split it to a token set $\mathcal Z=\{x_1,x_2,...,x_N\}$. For every token $x_i$, the complementary set is $\mathcal C=\mathcal Z-\{x_i\}$. We define all subsets of $\mathcal C$ as set $\mathcal S$ and define subsets of $\mathcal C$ consisting of $k$ elements as set $\mathcal S_{k}$ .
For the MIM method, we set $p$ as the mask ratio. The MIM models the following conditional probability:
$$
 \mathop{\mathbf E}\limits_{s\in \mathcal S_{(1-p)\times N}}P(x_i|s)
$$
And the stochastic autoregressive method models the following conditional probability:
$$
 \mathop{\mathbf E}\limits_{s\in \mathcal S}P(x_i|s)= \int_0^1\mathop{\mathbf E}\limits_{s\in \mathcal S_{(1-p)\times N}}P(x_i|s) \, \mathrm{d}p
$$
So stochastic autoregressive method is equivalent to multiple masked image models with different mask ratios. We think our method models more conditional probabilities and dependencies between tokens, and has more potential to perform better.

\section{More discussion}\label{sec:More_discussion}

\textbf{Broader impact.} 
For the enrichment of visual autoregression tasks, our proposed SAIM framework has been proved to be effective.
This method employs a stochastic attention masking strategy to provide global contextual semantic information to the model.
Therefore, our work gives a new perspective on visual self-supervised techniques and establishes the foundation for the development of unified autoregressive models for images and text.

\textbf{Potential negative impact.} 
The performance of visual self-supervised approaches for unlabeled data is improved by our work, which could benefit many useful applications such as autonomous driving and security.
However, there are several controversial risks associated with the technology, such as the possibility of personal private data in unlabeled data.
This problem is also a common issue with today's visual self-supervision approach, and it is getting public attention.
No other potential negative effects were found in our work.

\end{document}